\documentclass[journal]{IEEEtran}
\usepackage{graphicx}
\usepackage{amsmath}
\usepackage{amsfonts}
\usepackage{picins}
\usepackage[ruled, vlined]{algorithm2e}
\usepackage{multirow}

\newtheorem{defi}{\noindent \textbf{Definition}}

\begin{document}

\title{Developing an Unsupervised Real-time Anomaly Detection Scheme for Time Series with Multi-seasonality}
\author{Wentai~Wu, \IEEEmembership{Student~Member,~IEEE}, Ligang~He, \IEEEmembership{Member,~IEEE}, Weiwei~Lin, Yi~Su, Yuhua~Cui, Carsten Maple, and Stephen~Jarvis, \IEEEmembership{Member,~IEEE}
\thanks{Corresponding author: Ligang He. W. Wu, L. He and Y. Su are with the Department of Computer Science, University of Warwick. W. Lin is with the School of Computer Science and Engineering at the South China University of Technology. Y Cui is with the Research Institute of Worldwide Byte Information Security. C. Maple is with the Warwick Manufacturer Group, University of Warwick. S. Jarvis is with the College of Engineering and Physical Sciences, University of Birmingham.}}

\maketitle

\begin{abstract}
On-line detection of anomalies in time series is a key technique used in various event-sensitive scenarios such as robotic system monitoring, smart sensor networks and data center security. However, the increasing diversity of data sources and the variety of demands make this task more challenging than ever. Firstly, the rapid increase in unlabeled data means supervised learning is becoming less suitable in many cases. Secondly, a large portion of time series data have complex seasonality features. Thirdly, on-line anomaly detection needs to be fast and reliable. In light of this, we have developed a prediction-driven, unsupervised anomaly detection scheme, which adopts a backbone model combining the decomposition and the inference of time series data. Further, we propose a novel metric, Local Trend Inconsistency (LTI), and an efficient detection algorithm that computes LTI in a real-time manner and scores each data point robustly in terms of its probability of being anomalous. We have conducted extensive experimentation to evaluate our algorithm with several datasets from both public repositories and production environments. The experimental results show that our scheme outperforms existing representative anomaly detection algorithms in terms of the commonly used metric, Area Under Curve (AUC), while achieving the desired efficiency.
\end{abstract}


\section{Introduction}
Time series data sources have been of interest in a vast variety of areas for many years -- the nature of time series data was examined in a seminal study by Yule \cite{Yule1926} and the techniques were applied to areas such as econometric \cite{frisch1933} and oceanographic data \cite{seiwell1949} since the 1930s. However, in an era of hyperconnectivity, big data and machine intelligence, new technical scenarios are emerging such as autonomous driving, edge computing and Internet of Things (IoT).  Analysis of such systems poses new challenges to the detection of anomalies in time series data. Further, for a wide range of systems which require 24/7 monitoring services, it has become crucial to have the detection techniques that can provide early, reliable reports of anomalies. In cloud data centers, for example, a distributed monitoring system usually collects a variety of log data from the virtual machine level to the cluster level on a regular basis and sends them to a central detection module, which then analyzes the aggregated time series to detect any anomalous events including hardware failures, unavailability of services and cyber attacks. This requires a reliable on-line detector with strong sensitivity and specificity. Otherwise, the inefficient detection may cause unnecessary maintenance costs.

Several classes of schemes have been applied to the problem of anomaly detection for time series data. In certain cases decent results can be achieved by these traditional methods such as outlier detection \cite{ocsvm}\cite{isoforest}\cite{antihub}\cite{isoDC}, pattern (segment) extraction \cite{shapelet1999}\cite{shapelet2012}\cite{shapelet2018}\cite{shapelet2016} and sequence mapping \cite{SAX2005}\cite{PCA2015}\cite{FCM2017}.
However, we are facing a growing number of new scenarios and applications which produce large volumes of time series data with unprecedented complexity, posing challenges that traditional anomaly detection methods cannot address effectively. First, more and more time series data are being produced without labels since data labeling/annotation is usually very time-consuming and costly. Sometimes it is also unrealistic or impossible to acquire reliable labels when their correctness has to be guaranteed. Second, some applications may produce multi-channel series with complex features such as multi-period seasonality (i.e., multiple seasonal, such as yearly or monthly, patterns within one channel), long periodicity, fairly unpredictable channels and different seasonality between channels. As a result, learning these patterns requires effective seasonality discovery and strong ability of generalization. Third, the process is commonly required to be fast enough to support instant reporting or alarming once unexpected situation occurs. The capability of on-line detection is especially important in a wide range of event-sensitive scenarios such as medical and industrial process control systems. 

In this paper, we propose a predictive solution to detecting anomalies effectively in time series with complex seasonality. The fundamental idea is to inspect the data samples as they arrive and match the data samples with an ensemble of forecasts made chronologically. Specifically, our solution comprises an augmented forecasting model and a novel detection algorithm that exploits the predictions of local sequences made by the underlying forecasting model. We built a frame-to-sequence Gated Recurrent Unit (GRU) network while extending its input with seasonal terms extracted by decomposing the time series of each sample channel. The integration of the seasonal features can alleviate negative impact from anomalous samples in the training data since the anomalous samples have minor impact on the long-term periodic patterns. Because of the above reasons, our prediction framework does not require the labels (specifying which data are normal or abnormal) or uncontaminated training data (i.e., our solution tolerates polluted/abnormal training samples). 

After predicting local sequences (i.e., the output of the forecasting model), we use a novel method to weight the ensemble of different forecasts based on the reliability of their forecast sources and make it a chronological process to fit the on-line detection. The weight of each forecast is determined dynamically during the process of detection by scoring each forecast source (i.e., the forecast made based on this data source), which reflects how likely the predictions made by a forecast source is trustworthy. Based on the above ensemble, we propose a new metric, termed Local Trend Inconsistency (LTI), for measuring the deviation of an actual sequence from the predictions in real-time, and assigns an anomaly score to each of the newly arrived data points (which we also call frames) in order to quantify the probability that a frame is anomalous. 

We also propose a method to map the LTI value of a frame to its Anomaly Score (AS) by a logistic-shaped function. The mapping further differentiates anomalies and normal data. In order to determine the logistic mapping function, we propose a method to automatically determine the optimal values of the fitting parameters in the logistic mapping function. The AS value of a frame in turn becomes the weight of its impact on the detection of future frames. This makes our LTI metric robust to the anomalous frames in the course of detection and significantly mitigates the potential impact of anomalous samples on the detection results of the future frames. This feature also enables our algorithm to work chronologically without maintaining a large reference database or caching too many historical data frames. To the best of our knowledge, the existing prediction-driven detection schemes do not take into account the reliability of the forecast sources.

The main contributions of our work are as follows:
\begin{itemize}
\item We designed a frame-to-sequence forecasting model integrating a GRU network with time series decomposition (using Prophet, an additive time series model developed by Facebook \cite{Prophet_paper}) to enable the contamination-tolerant training on multi-seasonal time series data without any labels. 
\item We propose a new metric termed Local Trend Inconsistency (LTI), and based on this metric we further propose an unsupervised detection algorithm to score the probability of data anomaly. An practical method is also proposed for fitting the scoring function. 
\item{We mathematically present the computation of LTI in the form of matrix operations and prove the possibility of parallelization for further speeding up the detection procedure.}
\item We conducted extensive experiments to evaluate the proposed scheme on two public datasets from the UCI data repository and a more complex dataset from a production environment. The result shows that our solution outperforms the existing algorithms significantly with low detection overhead. 
\end{itemize}

The rest of this paper is organized as follows: Section II discusses a number of studies related to anomaly detection. In Section III, we introduce Local Trend Inconsistency as the key metric in our unsupervised anomaly detection scheme. We then systematically present our unsupervised anomaly detection solution in Section IV, including the backbone model for prediction and a scoring algorithm for anomaly detection. We present and analyze the experimental results in Section V, and finally conclude this paper in Section VI.

\section{Related Work}
\label{sec:related_work}
The term \textit{anomaly} refers to a data point that significantly deviates from the rest of the data which are assumed to follow some distribution or pattern. There are two main categories of approaches for anomaly detection: novelty detection and outlier detection. While novelty detection (e.g. classification methods \cite{classification1}\cite{classification2}\cite{classification3}\cite{classification4}) requires the training data to be classified, outlier detection (e.g., clustering, principal component analysis \cite{PCA2015} and feature mapping methods \cite{VCM}\cite{feature_learning}) does not need a prior knowledge of classes (i.e., labels) and thus is also known as unsupervised anomaly detection. The precise terminology and definitions of these terminology may vary in different sources. We use the same taxonomy as Ahmed et al. did in reference \cite{survey2016} whilst in the survey presented by Hodge and Austin \cite{survey2004} unsupervised detection is classified as a subtype of outlier detection. The focus of our work is on unsupervised anomaly detection since we aim to design a more generic scheme and thus do not need to assume the labels are unavailable. 

In the detection of time series anomalies, we are interested in discovering abnormal, unusual or unexpected records. In a time series, an anomaly can be detected within the scope of a single record or as a subsequence/pattern. Many classical algorithms can be applied to detect single-record anomaly as an outlier, such as the One Class Support Vector Machine (OCSVM) \cite{ocsvm}, a variant of SVM that exploits a hyperplane to separate normal and anomalous data points. Zhang et al. \cite{ocsvm1} implemented a network performance anomaly detector using OCSVM with Radial Basic Function (RBF), which is a commonly used kernel for SVM. Maglaras and Jiang \cite{ocsvm2} developed an intrusion detection module based on K-OCSVM, the core of which is an algorithm that performs K-means clustering iteratively on detected anomalies. Shang et al. \cite{ocsvm3} applies Particle Swarm Optimization (PSO) to find the optimal parameters for OCSVM, which they applied to detect the abnormalities in TCP traffic. In addition, Radovanovi\'c et al. \cite{antihub} investigated the correlation between hub points and outliers, providing a useful guidance on using reverse nearest-neighbor counts to detect anomalies. Liu et al. \cite{isoforest} found that anomalies are susceptible to the property of "isolation" and thus proposed Isolation Forest (iForest), an anomaly detection algorithm based on the structure of random forest. Taking advantage of iForest's flexibility, Calheiros et al. \cite{isoDC} adapted it to dynamic failures detection in large-scale data centers. For anomalous sequence or pattern detection, there are a number of classical methods available such as box modeling \cite{box}, symbolic sequence matching \cite{SAX2005} and pattern extraction \cite{shapelet2018}\cite{shapelet2016}). For example, Huang et al. \cite{LOF+SAX+cloud} proposed a scheme to identify the anomalies in VM live migrations by combining the extended Local Outlier Factor (LOF) and Symbolic Aggregate ApproXimation (SAX). 

Recent advance in machine learning techniques inspires prediction-driven solutions for intelligent surveillance and detection systems (e.g., \cite{surveil}\cite{protein}). A prediction-driven anomaly detection scheme is often a sliding window-based scheme, in which future data values are predicted and then the predictions are compared against the actual values when the data arrive. This type of anomaly detection schemes has been attracting much attention recently thanks to the remarkable performance of recurrent neural networks (RNNs) in prediction/forecasting tasks. Filonov et al. \cite{lstm-fd} proposed a fault detection framework that relies on a Long Short Term Memory (LSTM) network to make predictions. The set of predictions along with the measured values of data are then used to compute error distribution, based on which anomalies are detected. Similar methodologies are used by \cite{lstm-ad} and \cite{HTM}. LSTM-AD \cite{lstm-ad} is also a prediction scheme based on multiple forecasts. In LSTM-AD the abnormality of data samples is evaluated by analyzing the prediction error and the corresponding probability in the context of an estimated Gaussian error distribution obtained from the training data. However, the drawback of LSTM-AD is that it is prone to the contamination of training data. Therefore, when the training data contains both normal and anomalous data, the accuracy of the prediction model is likely to be affected, which consequently make the anomaly detection less reliable. 

Malhotra et al. \cite{encoder-decoder} adopt a different architecture named encoder-decoder, which is based on the notion that only normal sequences can be reconstructed by a well-trained encoder-decoder network. A major limitation of their model is that an unpolluted training set must be provided. As revealed by Pascanu et al. \cite{RNN_difficulty}, RNNs may struggle in learning complex seasonal patterns in time series particularly when some channels of the series have long periodicity (e.g., monthly and yearly). A possible solution to that is decomposing the series before feeding into the network. Shi et al. \cite{wavelet+BPNN} proposed a wavelet-BP (Back Propagation) neural network model for predicting the wind power. They decompose the input time series into the frequency components using the wavelet transform and build a prediction network for each of them. To forecast time series with complex seasonality, De Livera et al. \cite{Livera2011} adopt a novel state space modeling framework that incorporates the seasonal decomposition methods such as the Fourier representation. A similar model was implemented by Gould et al. \cite{Gould2008} to fit hourly and daily patterns in utility loads and traffic flows data.

Ensuring low overhead is essential for real-time anomaly detection. For example, Gu et al. \cite{gu2017} proposed an efficient motif (frequently repeated patterns) discovery framework incorporating an improved SAX indexing method as well as a trivial match skipping algorithm. Their experimental results on the CPU host load series show excellent time efficiency. Zhu et al. \cite{gu2018} propose a new method for locating similar sub-sequences as well as a parallel approach using GPUs to accelerate Dynamic Time Warping (DTW) for time series pattern discovery. Similarly, parallel algorithms (e.g., \cite{prll_cnn}\cite{multiclass}\cite{flink}) have been applied to several forms of machine learning models for efficiency boost.

\section{Local Trend Inconsistency}
\label{sec_III}
In this section, we first introduce a series of basic notions and frequently-used symbols, then define a couple of distance metrics, and finally present the core concept in our anomaly detection scheme -\emph{ Local Trend Inconsistency} (\emph{LTI}).

In some systems, more than one data collection device is deployed to gather information from multiple variables relating to a common entity simultaneously, which consequently generates multi-variate time series. In this paper we call them multi-channel time series. 
\begin{defi}  
A \textit{channel} is the full-length sequence of a single variable that comprises the feature space of a time series.
\end{defi}

For the sake of convenience, we define a frame as follows. This concept of a frame is inspired by, but is more general than, a frame in video processing (since a video clip can be reckoned as a time series of images.)
\begin{defi}  
A \textit{frame} is the data record at a particular point of time in a series. A frame is a vector in a multi-channel time series, or a scalar value in a single-channel time series.
\end{defi}

Most of previous schemes detect anomalies by analyzing the data items in a time series as separate frames. However, in our approach we attempt to conduct the analysis from the perspective of local sequences.
\begin{defi}  
A \textit{local sequence} is a fragment of the target time series; a local sequence at frame $x$ is defined as a fragment of the series spanning from a previous frame to frame $x$.
\end{defi}

For clarity, we list all the symbols frequently used in this paper in Table \ref{Tab: symbols}.
\begin{table}[ht]
\centering
\caption{List of symbols}
\begin{tabular}{ l l } 
 \hline
 Symbol 	  	 & Description \\ 
 \hline
 $X$			 	& A time series $X$ \\
 $X(t)$			 	& The $t$-th frame of time series $X$  \\
 $X^{(c)}$			& The $c$-th channel of time series $X$ \\
 $X^{(c)}(t)$		& The $c$-th component of the $t$-th frame of time series $X$ \\
 $x^{(i)}$		 	& The $i$-th feature of frame $x$\\
 $\hat{x}_k$	 	& The forecast of the frame $x$ predicted by frame $k$\\
 $S$			 	& An actual local sequence from the target time series \\
 $S_k$  	  	 	& A local sequence predicted by frame $k$ \\
 $S(i)$  	  	 	& The $i$-th frame in local sequence $S$ \\
 $S(i,j)$  	  	 	& An actual local sequence spanning from frame $i$ to $j$ \\
 $S_k(i,j)$   	 	& A local sequence predicted by $k$ spanning from frame $i$ to $j$ \\ 
 \hline
\end{tabular}
\label{Tab: symbols}
\end{table}

Euclidean Distance and Dynamic Time Warping (DTW) Distance are commonly used to measure the distance between two vectors. However, the scale of Euclidean Distance largely depends on the dimensionality, i.e., vector length. DTW distance can measure the sequence similarity, but cannot produce the length-independent results. With the relatively high time complexity ($O(n^2m)$ for $m$-dimensional sequences of length $n$), DTW is often applied to the sequence-level analysis, in which the target is a sequence of frames or a pattern of varying length. However, our work aims to perform the frame-wise, on-line detection, i.e., detect whether a frame is anomalous as the frame arrives. 

Therefore, in this paper we use a modified form of Euclidean distance, called Dimension-independent Frame Distance ($DFDist$) as formulated in Eq. (\ref{eq: DFDist}), to measure the distance between two frames $x$ and $y$:

\begin{equation}  
DFDist(x,y) = \frac{1}{m}\sum_{i=1}^{m} (x^{(i)} - y^{(i)})^2
\label{eq: DFDist}
\end{equation}
where $m$ is the number of dimensions (i.e., number of channels) and $x^{(i)}$ and $y^{(i)}$ are the $i$-th component of frame $x$ and frame $y$, respectively. We do not square root the result. This does not impact the effectiveness of our approach, but makes it easier to handle when we transform all computations into matrix operations at the later stage of the processing. Also, the desired scale (i.e., $DFDist \in [0,1]$) of the distance still holds for normalized data.

With $DFDist$, we can further measure the distance between two local sequences of the same length. The desired metric for sequence distance should be independent on the length of the sequences as we want to have a unified scale for any pair of sequences. We formulate the Length-independent Sequence Distance ($LSDist$) between two sequences $S_X$ and $S_Y$ of the same length in Eq. (\ref{eq: LSDist}), where $L$ is the length of the two local sequences.

\begin{equation}  
LSDist(S_X, S_Y) = \frac{1}{L}\sum_{i=1}^{L} DFDist(S_X(i), S_Y(i))
\label{eq: LSDist}
\end{equation} 

Although the definition of $LSDist$ already provides a unified scale of distance, the temporal information of the time series data is neglected. Assuming we are detecting the anomaly of the event at time $t$, we need to compare the local sequence at frame $t$ with a ground truth sequence (assume there is one) to see if anything goes wrong in the latest time window. If we use $LSDist$ as the metric, then every time point is regarded as being equally important. However, this does not practically comply with the rule of time decay, namely, the most recent data point typically has the greatest reference value and also the greatest impact on what will happen in the next time point. Therefore, we refine $LSDist$ by weighting each term and adding a normalization factor.
The Weighted Length-independent Sequence Distance ($WLSDist$) is defined in Eq. (\ref{eq: WLSDist}), where $d_i$ is the weight of time decay for frame $i$ and $D_L$ is the normalization factor (so that $WLSDist$ remains in the same scale as $LSDist$). 

\begin{equation}  
WLSDist(S_X, S_Y) = \frac{\sum_{i=1}^{L} d_i \cdot DFDist(S_X(i), S_Y(i))}{D_L}
\label{eq: WLSDist}
\end{equation}

Time decay is applied on the basis that the two sequences are chronologically aligned. In this paper, we use the exponentially decaying weights, which is similar to the exponential moving average method \cite{EMA}:
\begin{equation}  
d_i = e^{-(L-i)}, i=1,2,...,L
\label{eq: d_i}
\end{equation}
where $i$ denotes the frame index and $L-i$ is the temporal distance (with $i=L$ being the current frame). Hence, the corresponding normalization factor $D_L$ in Eq. (\ref{eq: WLSDist}) is the summation of a geometric series of length $L$:
\begin{equation}  
D_L = \sum_{i=1}^{L} e^{-(L-i)} \\
	= \frac{1-e^{-L}}{1-e^{-1}}
\label{eq: D_L}
\end{equation}
where $L$ is the sequence length.

Ideally it is easy to identify the anomalies by calculating $WLSDist$ between the target (such as local sequence or frame) and the ground truth. However, this approach is not feasible if the labels are unavailable (i.e., there is no ground truth). A possible solution is to replace the ground truth with expectation, which is obtained typically by using time series forecasting methods \cite{Chauhan2015}\cite{lstm-ad}, which is the basic idea of the so-called prediction-driven anomaly detection schemes. However, a critical problem with such a prediction-driven scheme is the reliability of forecast. On the one hand, the prediction error is inevitable. On the other hand, the predictions made based on the historical frames, which may include anomalous frames, can be unreliable. This poses a great challenge for prediction-driven anomaly detection schemes. 

Envisaging the above problems, we propose a novel, reliable prediction scheme, which makes use of multi-source forecasting. Unlike previous studies that use frame-to-frame predictors, our scheme makes a series of forecast at different time points (i.e, from different sources) by building a frame-to-sequence predictor. The resulting collection of forecasts form a \textit{common} expectation from multiple sources for the target. When the target arrives and if it deviates from the common expectation, it is deemed that the target is likely to be an anomaly. This is the underlying principle of our unsupervised anomaly detection. 

In order to quantitatively measure how far the target deviates from the collection of expectations obtained from multiple sources, we propose a metric we term the Local Trend Inconsistency (LTI). LTI takes into account the second challenging issue discussed above (i.e, there may exist anomalous frames in history) by weighting the prediction made based on a source (i.e., a frame at a previous time point) with the probability of the source being normal. 

For a frame $t$ (i.e., by which we refer to the frame arriving at time point \textit{t}), $LTI(t)$ is formally defined in Eq. (\ref{eq: LTI_prob}), where $S(i+1, t)$ is the actual sequence from frame $i+1$ to frame $t$, and $S_i(i+1, t)$ is the sequence of the same span predicted by frame $i$ (i.e., prediction made when frame $i$ arrives). $L$ is the length of the prediction window, which is a hyper-parameter determining the maximum length of the predicted sequence and also the number of sources that make the predictions (i.e., the number of predictions/expectations) of the same target. $P(i)$ denotes the probability of frame $i$ being normal. 

$Z_t$ is the normalization factor for frame $t$ defined as the sum of all the probabilistic weights shown in Eq. (\ref{eq: Z_i}). $Z_t$ is used to normalize the value of $LTI(t)$ to the range of $[0, 1]$.

\begin{equation}  
LTI(t) = \frac{1}{Z_t}\sum_{i=t-L}^{t-1} P(i) \cdot WLSDist\big(S(i+1,t), S_i(i+1,t)\big)
\label{eq: LTI_prob}
\end{equation}

\begin{equation}  
Z_t = \sum_{i=t-L}^{t-1} P(i)
\label{eq: Z_i}
\end{equation}

\begin{figure}[ht]
    \centering
    \includegraphics[width=3.2in]{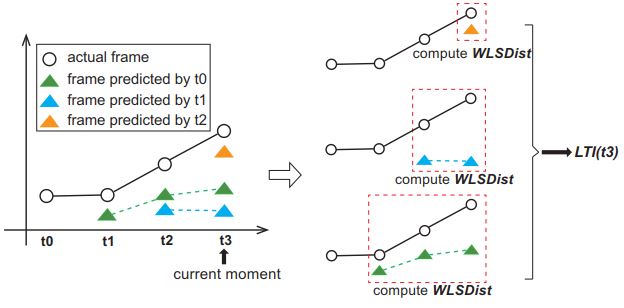}
    \caption{An example demonstrating the calculation of Local Trend Inconsistency with the max probe length $L$ equal to 3.}
    \label{fig:demo_LTI}
\end{figure}

Fig. \ref{fig:demo_LTI} illustrates how LTI is calculated in a case where $L=3$ (i.e., the length of the prediction window is 3). Based on the actual data arriving at $t_0$ (the actual data are represented by circles), our scheme predicts the frames at three future time points, i.e., $t_1$, $t_2$ and $t_3$, which are depicted as green triangles in the left part of Fig. \ref{fig:demo_LTI}. When the time elapses to $t_1$, the data at $t_1$ arrives and our scheme predicts the data at the time points of $t_2$, $t_3$ and $t_4$ (in the figure we only plot the predictions up to the time point $t_3$), which are colored blue. Similarly, when the time elapses to $t_2$, the data at $t_2$ arrives and our scheme forecasts the data at the time points of $t_3$, $t_4$ and $t_5$ (colored orange).  

Now assume we want to calculate $LTI(t_3)$ to gauge the abnormality of the data arriving at time $t_3$. As shown in Fig. \ref{fig:demo_LTI}, at time $t_3$, we know the actual local sequence from $t_0$ to $t_3$, i.e., $S(t_0,t_3)$ (corresponding to the term $S(i+1, t)$ in Eq. \ref{eq: LTI_prob}), and also we have made the following three predictions, which are the forecasts at three different time points:

\begin{itemize}
\item $S_0(t_1,t_3)$: the predicted local sequence from $t_1$ to $t_3$, which is predicted at time $t_0$;
\item $S_1(t_2,t_3)$: the predicted local sequence from $t_2$ to $t_3$, which is predicted at time $t_1$;
\item $S_2(t_3)$: the prediction of frame $t_3$ made at time $t_2$. 
\end{itemize}

$LTI(t_3)$ is then obtained by i) calculating the weighted distances (i.e., $WSLDist$ in Eq. \ref{eq: WLSDist}) between the predicted sequences and the corresponding actual sequence up to time $t_3$, i.e., the distances between $S_0(t_1,t_3)$ and $S(t_1,t_3)$ (shown at the bottom right of Fig. \ref{fig:demo_LTI}), between $S_1(t_2,t_3)$ and $S(t_2,t_3)$ (middle right of Fig. \ref{fig:demo_LTI}), and between $S_2(t_3)$ and $S(t_3)$ (top right of Fig. \ref{fig:demo_LTI}); ii) calculating the weighted sum (the weight is $P(i)$) of the distances obtained in last step, and iii) normalizing the weighted sum (i.e. divided by $Z_t$ in Eq. (\ref{eq: Z_i})). 

This multi-source prediction establishes the common expectation for the data values. How far the actual data deviates from the predicted data, which is measured by the distance between them, is used to quantify the abnormality of the given data.  

The whole process can be formulated using matrix operations. Assume we are detecting anomaly at frame $t$ and the size of the prediction window is $L$. For brevity let $df_k(t)$ denote the distance between frame $t$ and a forecast of the frame made at time $k$ (i.e., $DFDist(t,\hat{t}_k)$). We first define the frame-distance matrix $\mathbf{D_F}$:
$$
\mathbf{D_F} = 
\left[
\begin{array}{c}
  \mathbf{D_F}^{(t-L)} \\
  \mathbf{D_F}^{(t-L+1)} \\
  \vdots \\
  \mathbf{D_F}^{(t-1)} \\	
\end{array}
\right]
$$
where
$$
\mathbf{D_F}^{(u)} =
\left[
\begin{array}{c}
  df_{u}(u+1)	\\ 
  df_{u}(u+2)	\\ 
  \vdots			\\ 
  df_{u}(t)		\\	
\end{array}
\right]^\mathrm{T}
$$
Then we define two diagonal normalization matrices $\mathbf{N_1}$ and $\mathbf{N_2}$ as follows:
$$
\mathbf{N_1} =
\left[
\begin{array}{c c c c}
  \frac{1}{D_L}	&  						&  		& 0					\\ 
   				& \frac{1}{D_{L-1}} 	&  		&  					\\  	
  				& 		 				& \ddots&  					\\
  0				& 		 				& 		& \frac{1}{D_1} 	\\
\end{array}
\right]
$$
$$
\mathbf{N_2} =
\left[
\begin{array}{c c c c}
  \frac{1}{Z_t}	&  						&  		& 0					\\ 
   				& \frac{1}{Z_t} 		&  		&  					\\  	
  				& 		 				& \ddots&  					\\
  0				& 		 				& 		& \frac{1}{Z_t} 	\\
\end{array}
\right]
$$
where $D_L$ and $Z_t$ are defined in (\ref{eq: D_L}) and (\ref{eq: Z_i}), respectively. For convenience let $ds_k(t)$ denote $WLSDist\big(S(k+1,t),S_{k}(k+1,t)\big)$. Hence we can derive the matrix of weighted local sequence distances denoted as $\mathbf{D_S}$:
$$
\mathbf{D_S} =
\left[
\begin{array}{c}
  ds_{t-L}(t)	\\
  ds_{t-L+1}(t)	\\
  \vdots		\\
  ds_{t-1}(t)	\\
\end{array}
\right]
 = \mathbf{N_1} \mathbf{D_F} \mathbf{T}
$$
where $\mathbf{T}$ is the time decay vector defined as:
$$
\mathbf{T} =
\left[
\begin{array}{c}
  d_1	\\
  d_2	\\
  \vdots\\
  d_L	\\
\end{array}
\right]
$$
where $d_i$ is computed via Eq. (\ref{eq: d_i}). Now we assume the probability of being normal is already known for each of frame $t$'s predecessors (i.e., $P(t-1), P(t-2),...$), and we put them together into a $1 \times L$ matrix $\mathbf{P}$:
$$
\mathbf{P} =
\left[
\begin{array}{c c c c}
  P(t-L) & P(t-L+1) & \cdots & P(t-1)	\\
\end{array}
\right]
$$
Then we can reformulate $LTI(t)$ as below:
\begin{equation}
LTI(t) = \mathbf{P}\mathbf{N_2}\mathbf{D_S}
	   = \mathbf{P}\mathbf{N_2}\mathbf{N_1}\mathbf{D_F}\mathbf{T}
\label{eq:LTI_matrix}
\end{equation}

Through the use of matrices to formulate the calculation of $LTI$, we can know that the calculation can be performed efficiently in parallel. The Degree of Parallelism (DoP) of its calculation can be higher than $L$. This is because the DoP for calculating the \textit{L} terms in Eq. (\ref{eq: LTI_prob}) can be $L$ apparently (the calculation of every term is independent on each other). The calculation of each term can be further accelerated (including the calculations of $WLSDist$ and $DFDist$) by parallelizing the matrix multiplication. For example, with a number of $L \times L$ processes (i.e., a grid of processes) and exploiting the Scalable Universal Matrix Multiplication Algorithm (SUMMA) \cite{SUMMA}, we can achieve a roughly $L^2$ speedup in the multiplication of any two matrices with the dimension size of $L$, which helps reduce the time complexity of computing $\mathbf{N_1}\mathbf{D_F}$ from $O(L^3)$ to $O(L)$. Further, with the resulting $\mathbf{N_1}\mathbf{D_F}$ the computation of $\mathbf{N_1}\mathbf{D_F}\mathbf{T}$ and $\mathbf{P}\mathbf{N_2}$ can be performed in parallel as both of them are vector-matrix multiplication requiring only $L$ processes and have time complexity of $O(L^2/L)=O(L)$. Finally multiplying the resulting matrices of $\mathbf{P}\mathbf{N_2}$ (dimension=$1\times L$) and $\mathbf{N_1}\mathbf{D_F}\mathbf{T}$ (dimension=$L\times 1$) consumes $O(L)$. Note that the matrix $D_F$ contains $L\times L$ entries of frame distance, each of which is calculated using Eq. (\ref{eq: DFDist}). Therefore, updating $D_F$ (upon a new frame arrives) is an operation with the complexity of $O(L^2m/L^2)=O(m)$, where $m$ is the frame dimension. Consequently, the time complexity of computing $LTI(t)$ in parallel is $O(m+L)$ in theory.

\section{Anomaly Detection with LTI}
Our anomaly detection scheme is based on LTI (Local Trend Inconsistency) as LTI can effectively indicate how significantly the series deviates locally from the common expectation established by multi-source prediction. 

As can be seen from Eq. (\ref{eq: LTI_prob}), there are still two problems to be solved in calculating $LTI$. First, a mechanism is required to make reliable predictions of local sequences. Second, we need an algorithm to quantify the probabilistic factors (in matrix $\mathbf{P}$) as they are not known apriori. 

In this section, we first introduce the backbone model we build for achieving accurate frame-to-sequence forecasting. The model is designed to learn the complex patterns in multi-seasonal time series with tolerance to pollution in the training data. Then we illustrate how to make use of the predictions (from multiple source frames) made to compute LTI. Finally, we propose an anomaly scoring algorithm that uses a scoring function to chronologically calculate anomaly probability for each frame based on LTI.

\subsection{Prediction Model}
To effectively learn and accurately predict local sequences in multi-seasonal time series, we adopt a combinatorial backbone model composed of a decomposition module and an inference module. 

Recurrent Neural Network (RNN) is an ideal network to implement the inference module of our prediction model. RNNs (including mutations such as Long Short Term Memory (LSTM) and Gated Recurrent Unit (GRU)) are usually applied as end-to-end models (e.g., \cite{two-dim_LSTM2019} \cite{end2end_LSTM&GRU}). However, a major limitation of them is the difficulty in learning complex seasonal patterns in multi-seasonal time series. Even though the accuracy may be improved by stacking more hidden layers and increasing back propagation distance (through time) during training, it could cause prohibitive training cost. 

In view of this, we propose to include the seasonal features of the input data explicitly as the input of the neural network. This is achieved by conducting time series decomposition before running the prediction model, which is the purpose of the decomposition module. The resulting seasonal features can be regarded as the outcome of feature engineering. Technically speaking, seasonal features are essentially the "seasonal terms" decomposed from each channel of the target time series. We use \emph{Prophet} \cite{Prophet_paper}, a framework based on the decomposable time series model \cite{Harvey_and_Peters1990}, to extract the channel-wise seasonal terms. Let $X^{(c)}$ denote the $c$-th channel of time series $X$, and $X^{(c)}(t)$ the $t$-th record of the channel. The outcome of time series decomposition for channel $c$ is formulated as below:
\begin{equation}
\label{eq1} 
    X^{(c)}(t) = g_c(t) + s_c(t) + h_c(t) + \epsilon
\end{equation}
where $g_c(t)$ is the trend term that models non-periodic changes, $s_c(t)$ represents the seasonal term that quantifies the seasonal effects. $h_c(t)$ reflects the effects of special occasions such as holidays, and $\epsilon$ is the error term that is not accommodated by the model. For simplicity, we in this paper only consider daily and weekly seasonal terms as additional features for the inference module of our model. \emph{Prophet} relies on Fourier series to model multi-period seasonality, which enables the flexible approximation of any periodic patterns with arbitrary length. The underlying details can be referred to \cite{Prophet_paper}.

Separating seasonal terms from original frame values and using them as additional features effectively improve RNN from the following perspectives. First, explicit input of seasonal terms helps reduce the difficulty of learning complex seasonal terms in RNN. The extracted seasonal terms quantify seasonal effects. Second, time cost of training is expected to decrease as we can apply the Truncated Back Propagation Through Time (TBPTT) with a distance much shorter than the length of periodicity. Besides, the series decomposition process is very efficient, which will be demonstrated later by experiments. The top part of Fig. \ref{fig:model_view} shows the architecture of our backbone prediction model. In the prediction model, a stacked GRU network is implemented as the inference module, which takes as input the raw features of a frame concatenated with its seasonality features. We demonstrate the effectiveness of this backbone model in Section \ref{subsec_exp1}.

\subsection{Computing LTI based on Predictions}
\begin{figure*}[ht]
    \centering
    \includegraphics[width=4.9in]{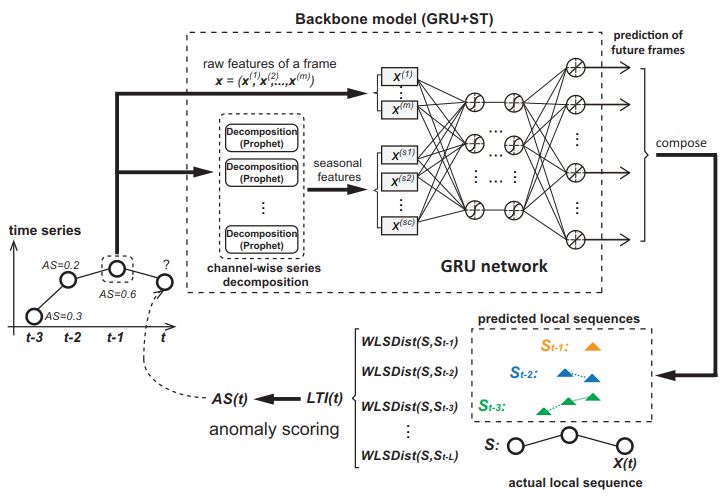}
    \caption{An overview of the proposed prediction-driven anomaly detection framework for the time series, which uses a seasonality augmented GRU network as the backbone model to support the abnormality scoring based on Local Trend Inconsistency (LTI).}
    \label{fig:model_view}
\end{figure*}

When we calculate Local Trend Inconsistency (LTI) in Eq. (\ref{eq: LTI_prob}), we are actually measuring the distance between a local sequence and an ensemble of its predictions by a well trained backbone prediction model. The workflow of our on-line anomaly detection method includes three main steps: i) feed every arriving frame into the prediction model and continuously gather its output of predicting future frames, ii) organize the frame predictions by their sources (i.e., the frames which made the forecast) and concatenate them into local sequences, and iii) compute LTI of the newly arrived frame according to Eq. (\ref{eq: LTI_prob}). Fig. \ref{fig:model_view} demonstrates the entire process, in which LTI of a frame is converted to a score of abnormality using the algorithms to be introduced later.

\subsection{Anomaly Scoring}
In theory, the values of $LTI(t)$ can be directly used to score frame $t$ in terms of its abnormality. However, the range of this metric is application-specific. So we further develop a measure that can represent the probability of data anomaly. Specifically, we define a logistic mapping function to convert the value of $LTI(t)$ to a probabilistic value:

\begin{equation}
 \Phi(x) = \frac{1}{1+e^{-k(x-x_0)}}
 \label{eq: Phi}
\end{equation}
where $k$ is the logistic growth rate and $x_0$ the x-value of the function's midpoint. 

\begin{figure}[ht]
    \centering
    \includegraphics[width=3.0in]{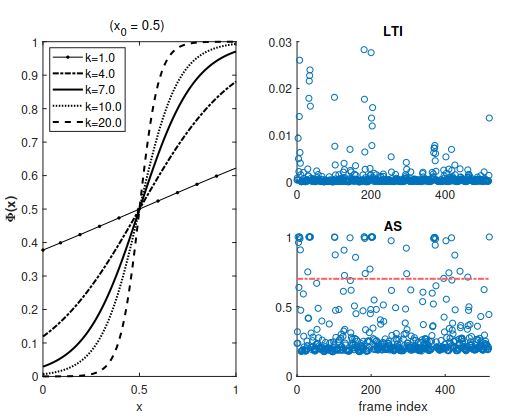}
    \caption{The mapping function $\Phi(\cdot)$ we use for anomaly scoring (left), and the dispersion effect by mapping $LTI$ values (top right) to anomaly scores (bottom right) with $\Phi(\cdot)$.}
    \label{fig:Phi_LTI_AS}
\end{figure}

The left part of Fig. \ref{fig:Phi_LTI_AS} shows the shapes of $\Phi(\cdot)$ with different values of $k$ when $x_0$ is set to 0.5. The shape of $\Phi(\cdot)$ becomes steeper as $k$ increases. We will introduce how to determine the optimal values of $k$ and $x_0$ later.

Now we define the probabilistic anomaly score of frame $t$ as below:
\begin{equation}
  AS(t) = \Phi(LTI(t))
  \label{eq: AS}
\end{equation}

The reason why we use Eq. (\ref{eq: Phi}) to map $LTI(t)$ to $AS(t)$ are three folds. First, we find that the $LTI(t)$ values are clustered together closely (top right of Fig. \ref{fig:Phi_LTI_AS}), which means that the difference in $LTI(t)$ values between normal and abnormal frames are not significant. This makes it difficult to differentiate them in practice although we can do so in theory. The right part of Fig. \ref{fig:Phi_LTI_AS} illustrates the situation where we map raw $LTI(t)$ values to $AS(t)$. It can be seen from the figure that the value of anomaly scores are better dispersed leaving a clearer divide between normal data and (potential) anomalies. For example, the red line we draw separates out roughly 10 percent of potential anomalies with high scores. Second, as discussed in the previous section, our scheme makes a series of forecast from different sources for the target, which establishes a common expectation for the target. The challenge is that there may exist anomalous sources, from which the forecast made is unreliable. Thus we have to differentiate the quality of the predictions by specifying large weights (i.e., the $P(i)$ in Eq. \ref{eq: LTI_prob}) for normal sources and small weights for the sources that are likely to be abnormal. With the function $\Phi(\cdot)$ to disperse the $LTI(t)$ values (by mapping them into $AS(t)$, the impact difference between normal and abnormal frames is magnified. Last but not the least, we find that the actual values of $LTI(t)$ depend on particular applications that our detection scheme is applied to. After mapping, the $AS(t)$ values becomes less application-dependent, making it possible to set a universal anomaly threshold. This is similar to the scenario of determining the unusual events if the samples follow the normal distribution: the values lying beyond two standard deviations from the mean are often regarded as unusual. 

Considering the second reason discussed above, we replace $P(i)$ in Eq. (\ref{eq: LTI_prob}) with $1-AS(i)$ where $i=t-L, t-L+1,..., t-1$. Consequently, $LTI(t)$ is reformulated as:
\begin{multline}
  LTI(t) = \\
  \frac{1}{Z_t}\sum_{i=t-L}^{t-1} (1-AS(i)) \cdot WLSDist\big(S(i+1,t), S_i(i+1,t)\big)
  \label{eq: LTI_AS}
\end{multline}
where $Z_t$ is the normalization factor reformulated as $\sum_{i=t-L}^{t-1}(1-AS(i))$ and $1-AS(i)$ represents the probability that frame $i$ is normal. 

The function $\Phi(\cdot)$ contains two parameters, $k$ and $x_0$.  The values of these two parameters need to be set before the function can be used to calculate the anomaly. Since $x_0$ is supposed to the midpoint of $x$, we set $x_0$ to be $\mathrm{mean}(LTI)$. We set $k$ to $c/\mathrm{stdev}(LTI)$ ($\mathrm{stdev}(LTI)$ is the standard deviation of $LTI$, and $c$ is a constant multiplier). The purpose of the mapping function is to disperse the LTI values that are densely clustered. On the one hand, the standard deviation $\mathrm{stdev}(LTI)$ can be used to represent how densely the LTI values reside around the mean. The lower the value of $\mathrm{stdev}(LTI)$, the more closely the LTI values are clustered. On the other hand, $k$ represents how steep the middle slope of the logistic mapping function is. The greater $k$ is, the steeper the logistic mapping function is. The more densely clustered the LTI values are, the steeper the logistic function needs to be in order to disperse those values. Therefore, for a set of LTI values with lower deviation, a bigger value should be set for $k$. 

Instead of setting the values of $k$ and $x_0$ manually, we propose an automated approach in this work to determine their values. More specifically, we design an iterative algorithm. The algorithm runs on a reference time series which is a portion of the training data. The algorithm is outlined in Algorithm \ref{algo: iterative}.

\begin{algorithm}[ht] 
\caption{Iterative procedure for unparameterizing $\Phi(\cdot)$}
 \DontPrintSemicolon
 \SetKwInOut{Input}{Input}
 \SetKwInOut{Output}{Output}
 \SetAlgoLined
 \Input{prediction span $L$, reference series length $r$, predicted local sequences $S_i(i+1, i+L)$ for $i \in [0,r-1]$}
 \Output{$k$, $x_0$}
 $k \gets 1.0$, $x_0 \gets 0.5$\;
 $AS(i) \gets 0$ for all $i \in [0,r-1]$\;
 \While{convergence criterion is not satisfied}{
   \For{$t \gets L$ \KwTo $r-1$}{
     compute $LTI(t)$ via Eq. (\ref{eq: LTI_AS})\;
     compute $AS(t)$ via Eq. (\ref{eq: AS})\;
   } 
   $k \gets \frac{c}{\mathrm{stdev}(LTI)}$, $x_0 \gets \mathrm{mean}(LTI)$\;
 }
\label{algo: iterative}
\end{algorithm}

In Algorithm \ref{algo: iterative}, parameters $k$ and $x_0$ are set to $1.0$ and $0.5$ initially, respectively. Note that it does not matter much what the initial values of $k$ and $x_0$ are. When Algorithm \ref{algo: iterative} is run on the reference time series, LTI for each frame of the reference series is calculated. The values of $k$ and $x_0$ will converge to $c/\mathrm{stdev}(LTI)$ and $\mathrm{mean}(LTI)$ eventually. In the algorithm, we set a convergence criterion, in which both $k$ and $x_0$ change by less than 0.1\% since last update. In each loop, the algorithm computes $LTI(t)$ and $AS(t)$ along the reference series for each frame $t$. After each loop, we update $k$ and $x_0$ and check if the criterion is met. 

With the anomaly scoring function $AS(\cdot)$ and the trained backbone model for the target series, we now present our Anomaly Detection based on Local Trend Inconsistency (AD-LTI). Assume we are detecting the anomaly for frame $t$, the pseudo-code of our on-line detection procedure is described in Algorithm \ref{algo: AD-LTI}.

\begin{algorithm}[ht] 
\caption{Anomaly Detection based on LTI}
 \DontPrintSemicolon
 \SetKwInOut{Input}{Input}
 \SetKwInOut{Output}{Output}
 \SetAlgoLined
 \Input{current frame $t$, prediction span $L$, previous frames from $t-L$ to $t-1$, $AS(i)$ for $i \in [t-L,t-1]$}
 \Output{$AS(t)$}
 \For{$i \gets t-L$ \KwTo $t-1$}{
   use the proposed prediction model to forecast $S_i(i+1,t)$\;
   compute $WLSDist\big(S(i+1,t),S_i(i+1,t)\big)$\;
 }
 compute $LTI(t)$ according to Eq. (\ref{eq: LTI_AS})\;
 compute $AS(t)$ according to Eq. (\ref{eq: AS})\;
\label{algo: AD-LTI}
\end{algorithm}

The information required for detection at frame $t$ includes frame $t$ itself, anomaly scores of previous frames, and the predicted local sequences ending at $t$, which is the output of our backbone prediction model. To analyze the time complexity of Algorithm \ref{algo: AD-LTI}, let $m$ denote the number of dimensions of a frame (i.e., channels of the time series) and $L$ the prediction span, which is a hyper-parameter shared by the backbone prediction model and the detection algorithm. Without parallelization, it takes $O(m)$ to calculate $DFDist$ between each pair of frames, so the time cost for obtaining $WLSDist$ between two local sequences is $O(Lm)$. Therefore, the time complexity of detection at a single frame $t$ is $O(L^2m)$ since $L$ sources of forecast are used (see Eq. \ref{eq: LTI_AS}). As analyzed in Section \ref{sec_III}, the complexity can be reduced to $O(m+L)$ with the proper parallelization.

\section{Experiments}
\label{sec_exp}
In this section, we first evaluate the effectiveness of our backbone prediction model. Then we compare AD-LTI with the existing anomaly detection algorithms in sensitivity and specificity (using the AUC metric). 

We set up our experiments on a machine equipped with a dual-core CPU (model: Intel Core i5-8500, 3.00 GHz), a GPU (model: GTX 1050Ti) and 32GB memory. The inference module of our backbone model is implemented on Pytorch (version: 1.0.1) platform and the decomposition module is implemented using Prophet (version: 0.4) released by Facebook. We select three datasets for evaluation. CalIt2 and Dodgers Loop Sensor are two public datasets published by the University of California Irving (UCI) and available in the UCI machine learning repository. Another dataset we use is from the private production environment of a cyber-security company, which is the collaborator of this project. This dataset collects the server logs from a number of clusters (owned by other third-party enterprises) on a regular basis. The dataset is referred to as the Server Log dataset in this paper.

\bigskip
\noindent \emph{CalIt2 Dataset}

CalIt2 is a multivariate time series dataset containing 10080 observations of two data streams corresponding to the counts of in-flow and out-flow of a building on UCI campus. The purpose is to detect the presence of an event such as a conference and seminar held in the building. The timestamps are contained in the dataset. The original data span across 15 weeks (2520 hours) and is half-hourly aggregated. We truncated the last 120 hours and conducted a simple processing on the remaining 2400 hours of data by making it hourly-aggregated. The CalIt2 dataset is provided with annotations that label the date, start time and end time of events over the entire period. There are 115 anomalous frames (4.56\% contamination ratio) in total. In our experiment, labels are omitted during training (because our prediction model forecasts local sequences of frames) and will only be used for evaluating detecting results. 


\bigskip
\noindent \emph{Server Log Dataset}

\begin{figure}[ht]
    \centering
    \includegraphics[width=3.0in]{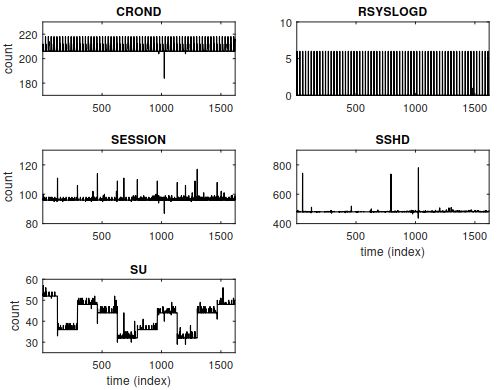}
    \caption{The Server Log time series dataset}
    \label{fig: serverlog_view}
\end{figure}

The Server Log dataset is a multi-channel time series with a fixed interval between two consecutive frames. The dataset spans from June 29th to September 4th, 2018 (1620 hours in total). The raw data is provided to us in form of separate log files, each of which stores the counts of a Linux server event on an hourly basis. The log files record the invocations of five different processes, which include CROND, RSYSLOGD, SESSION, SSHD and SU. Each process represents a channel of observing the server. We pre-processed the data by aggregating all the files to form a five-channel time series. Fig. \ref{fig: serverlog_view} shows the time series of all five channels. 

Currently, the company relies on security technicians to observe the time series and spot the potential anomalies, which might be caused by the security attacks. The aim of this project is to develop the automated method to spot the potential anomalies and quantify them at real time as the process invocations are being logged in the server. Anomalous events such as external cyber attacks exist in the Server Log dataset, but the labels are not available. We acquired the manual annotations for the test set from the technicians in the company. Totally 76 frames are labeled as anomalies in the test set, equivalent to a contamination ratio of 14.6\%.

\bigskip
\noindent \emph{Dodgers Loop Sensor Dataset}

Dodgers Loop Sensor is also a public dataset available in the UCI data repository. The data were collected at the Glendale on-ramp for the 101 North freeway in Los Angeles. The sensor is close enough to the stadium for detecting unusual traffic after a Dodgers game, but not so close and heavily used by the game traffic. Traffic observations were taken over 25 weeks (from Apr. 10 to Oct. 01, 2005) with date and timestamps provided for both data records and events (i.e., the start and end time of games). The raw dataset contains 50400 records in total. We pro-processed the data to make it an hourly time series dataset.

\subsection{Evaluating Backbone Model}
\label{subsec_exp1}
We trained our prediction model on the datasets separately to evaluate its accuracy as well as the impact of seasonal terms extracted by the decomposition module. We split the datasets into training, validation and test sets. On CalIt2, the first 1900 frames were used for training and the following 500 for testing. On the Server Log dataset, 1100 frames for training and 520 for test. On Dodgers Loop, 3000 records for training and 1000 for test. 300, 300, and 500 frames were used for validation on CalIt2, Server Log and Dodgers Loop, respectively. 

\begin{figure}[ht]
    \centering
    \includegraphics[width=3.0in]{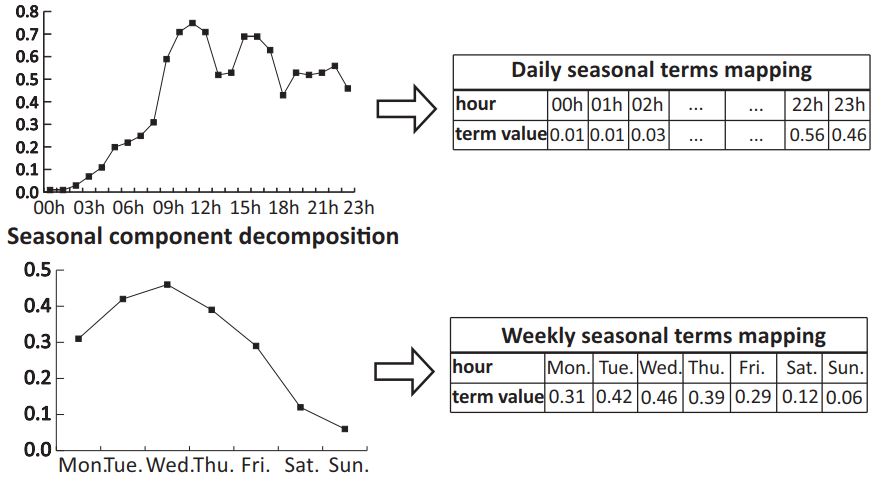}
    \caption{An example of seasonal terms mapping in which the numerical values quantify seasonal impacts}
    \label{fig:seasonal_terms}
\end{figure}

The proposed model uses \emph{Prophet} to implement the decomposition module and a stacked GRU network to implement the prediction module. We extracted daily and weekly terms for each channel. More specifically, for each channel we generated two mapping lists after fitting the data by \emph{Prophet}. One list contains the readings at each of 24 hours in a day, while the other list includes the readings at each of 7 days in a week. Fig. \ref{fig:seasonal_terms} shows an example of the mapping lists.

The values of seasonal terms are different for CalIt2, Server Log and Dodgers datasets, but the resulting mapping lists share the same format as the example shown in Fig. \ref{fig:seasonal_terms}. 

Based on the mapping lists and the timestamp field provided in the data we build our prediction network with seasonal features as additional input. Table \ref{Tab: network_topo} shows the network structures adopted for each of the datasets, where $L$ is the maximum length of local sequences as a hyper-parameter. $tanh$ is used as the activation function and Mean Square Error (MSE) loss as the loss function. Dropout is not enabled and we set a weight decay of $6e-6$ during the training to prevent over-fitting. We use Adam \cite{Adam} as the optimizer with the initial learning rate set to 0.001.

\begin{table}[ht]
\caption{Network structures of the inference part for predicting local sequence of length $L$}
\centering
\begin{tabular}{ l c c c} 
 \hline
 Dataset 	& type	& \# of features (raw+seasonal)	& topology	\\
 \hline
 CalIt2		& GRU	& 2+4							& $[6, 20{\times}2, 2L]$ \\
 Server	log	& GRU	& 5+10							& $[15, 20{\times}3, 5L]$ \\
 Dodgers	& GRU 	& 1+2							& $[3, 20{\times}2, L]$  \\
 \hline
\end{tabular}
\label{Tab: network_topo}
\end{table}

In order to evaluate the impact of the concatenated seasonal features, we also implemented a baseline GRU network with the same structure and hyper-parameters as our inference module except that the seasonal features are not included. We also consider the impact of a critical hyper-parameter, \emph{time\_steps}, in training the inference networks. The larger the \emph{time\_steps}, the longer the gradients back-propagate through time and the more time-consuming the training process becomes. We set different values of \emph{time\_steps} for the training of both our inference network and the baseline network to investigate the impact of seasonal features. The prediction span $L$ is fixed to 5 (hours). The results are summarized in Table \ref{tab:step_size_comp1}.

\begin{table*}[ht]
\caption{Comparing GRU+ST (the proposed backbone model augmented with seasonal features) with the vanilla GRU in accuracy, which is indicted by the lowest test MSE (Mean Square Error) achieved under different training settings of $time\_steps$ ($ts$). In each group of comparison, both models have converged and trained for the same number of epochs.}
\centering
\begin{tabular}{c l l l l l l l } 
 \hline
 &   &\multicolumn{2}{c}{Calit2 Dataset}&\multicolumn{2}{c}{Server Log Dataset}&\multicolumn{2}{c}{Dodgers Loop Dataset}													\\ 
 &   						& GRU+ST 	& GRU 		& GRU+ST 	& GRU 	 & GRU+ST 	& GRU 	\\
 \hline
 &	seasonal term decomp. time			& 2.7s 		& -   		& 6.6s 		& - 	 & 2.9s		& -	 	\\
 \hline
 \multirow{2}{*}{$ts$=24}&	Test MSE 	& 0.0068	& 0.0092 	& 0.0020 	& 0.0039 & 0.0098	& 0.0113\\ 
 &	Training time to converge			& 173.7s	& 176.7s	& 460.4s 	& 464.9s & 550.2s	& 552.3s\\
 \hline
 \multirow{2}{*}{$ts$=72}&	Test MSE 	& 0.0066	& 0.0089 	& 0.0013	& 0.0020 & 0.0066	& 0.0085\\
 &	Training time to converge			& 180.2s	& 185.6s	& 468.3s 	& 436.9s & 628.6s	& 632.9s\\
 \hline
 \multirow{2}{*}{$ts$=168}&	Test MSE 	& 0.0067	& 0.0085 	& 0.0018	& 0.0033 & 0.0072	& 0.0086\\
 &	Training time to converge			& 169.8s	& 174.5s	& 421.0s 	& 446.5s & 709.9s	& 752.9s\\
 \hline
\end{tabular}
\label{tab:step_size_comp1}
\end{table*}

In Table \ref{tab:step_size_comp1}, the decomposition time and training time refer to the fitting/training time spent by the decomposition module and the inference module, respectively. We evaluated three cases where \emph{time\_steps} takes different values of 24 (daily seasonality length), 72 or 168 (weekly seasonality length). Mean squared error (MSE) is calculated on the normalized test data to reflect the model quality. From the results we can first observe that it only takes the decomposition module of our model a few seconds to extract the seasonal terms from all the channels. More importantly, we find that augmenting the GRU model with seasonal terms (ST) makes the backbone model (GRU+ST) more complicated in structure, but it does not increase the training cost while resulting in much better accuracy -- it outperforms the baseline GRU network (without Seasonal Terms) significantly in accuracy (i.e., lower error). The accuracy increases by more than 20 percent on CalIt2 and by from 35 to nearly 50 percent on the Server Log dataset.

\subsection{Evaluating AD-LTI}
\label{subsec_exp2}
In this section we evaluate our unsupervised anomaly detection algorithm AD-LTI. We also implement a number of representative related algorithms for comparison. These baseline algorithms include One Class Support Vector Machine (OCSVM) \cite{ocsvm}, Isolation Forest (iForest) \cite{isoforest}, Piecewise Median Anomaly Detection \cite{piecewise}, LSTM-based Fault Detection (LSTM-FD) \cite{lstm-fd} and LSTM-AD, which is LSTM-based anomaly detection scheme using multiple forecasts \cite{lstm-ad}. 

OCSVM is a mutation of SVM for unsupervised outlier detection. OCSVM shares the same theoretical basis as SVM while using an additional argument $\nu$ as an anomaly ratio-related parameter. Isolation forest is an outlier detection approach based on random forest in which isolation trees are built instead of decision trees. An a priori parameter $cr$ is required to indicate the contamination ratio. Both OCSVM and Isolation Forest are embedded in the Scikit-learn package \cite{sklearn}. Piecewise Median Anomaly Detection is a window-based algorithm that splits the series into fixed-size windows within which anomalies are detected based on a decomposable series model. LSTM-FD is a typical prediction-driven approach that detects anomalies in time series by simply analyzing (prediction) error distribution. They adopt a frame-to-frame LSTM network as their backbone model. Similar to our approach, LSTM-AD also uses a multi-source prediction scheme (we discussed its working in Section \ref{sec:related_work}).

We use the AUC metric to measure the effectiveness. Area Under the Curve, abbreviated as AUC, is a commonly used metric for comprehensively assessing the performance of binary classifiers. "The curve" refers to the Receiver Operator Characteristic (ROC) Curve, which is generated by plotting the true positive rate (y-axis) against the false positive rate (x-axis) based on the dynamics of decisions made by the target classifier (the anomaly detector in our case). The concept of ROC and AUC can reveal the effectiveness of a detection algorithm from the perspectives of both specificity and sensitivity. Another reason why we choose AUC is because it is a threshold-independent metric. AD-LTI does not perform classification but presents the detection results in the form of probability. Hence metrics such as precision and recall cannot be calculated unless we consider the threshold as an extra parameter, which violates our aim of designing a generic scheme.

We evaluate AD-LTI and the baseline algorithms on these three datasets. Parameters for baseline algorithms are set to the default or the same as in the original papers if they were suggested. For LSTM-FD, LSTM-AD and AD-LTI, $time\_steps$ is set to 72 (hours).

\begin{figure}[hbt]
    \centering
    \includegraphics[width=2.6in]{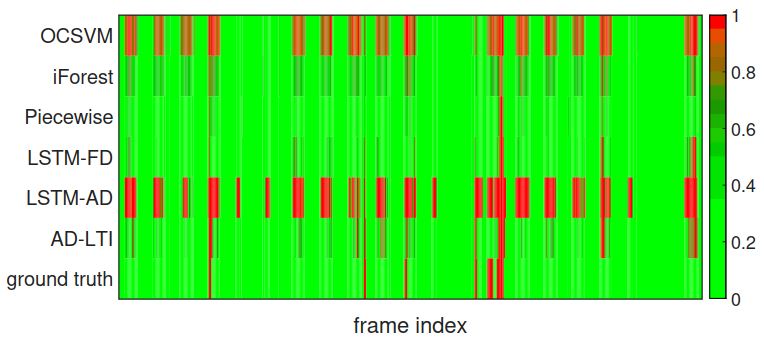}
    \caption{Heatmaps of detection decisions made by AD-LTI and baseline algorithms compared with the ground truth on CalIt2 dataset}
    \label{fig:heatmap_calit2}
\end{figure}

\begin{figure}[hbt]
    \centering
    \includegraphics[width=2.6in]{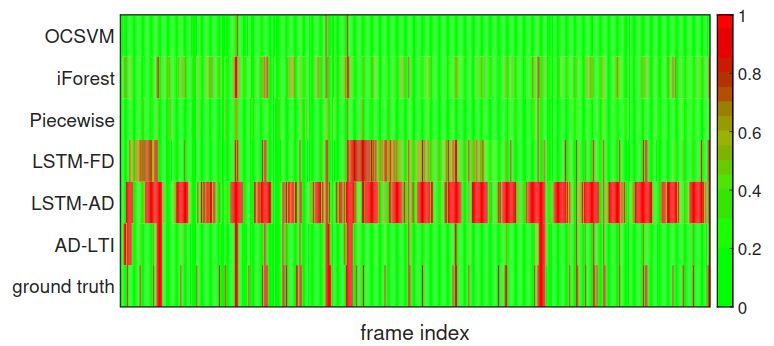}
    \caption{Heatmaps of detection decisions made by AD-LTI and baseline algorithms compared with the ground truth on Server Log dataset}
    \label{fig:heatmap_serverlog}
\end{figure}

\begin{figure}[hbt]
    \centering
    \includegraphics[width=2.6in]{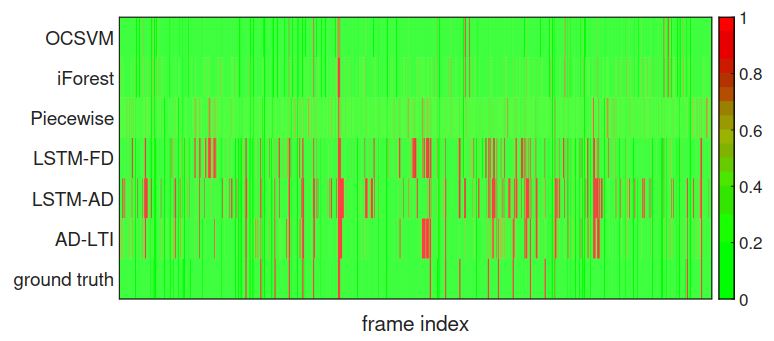}
    \caption{Heatmaps of decision results by AD-LTI and baseline algorithms compared with the ground truth on Dodgers Loop dataset}
    \label{fig:heatmap_dodgers}
\end{figure}

\begin{figure}[hbt]
    \centering
    \includegraphics[width=3.0in]{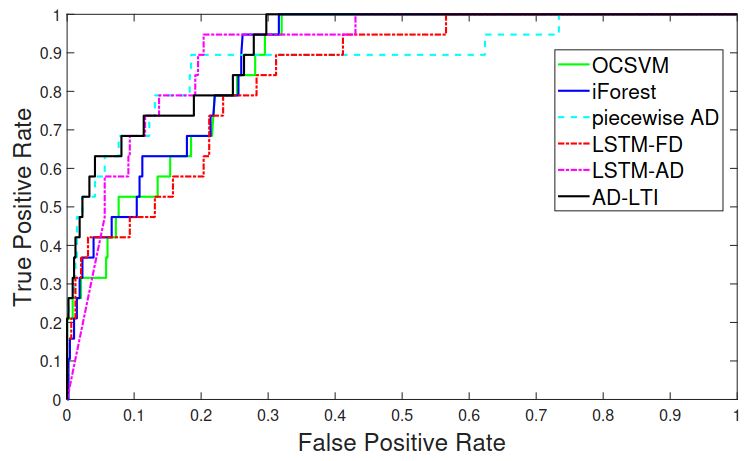}
    \caption{ROC curves of anomaly detection algorithms on Calit2 dataset}
    \label{fig:ROC_calit2}
\end{figure}

\begin{figure}[hbt]
    \centering
    \includegraphics[width=3.0in]{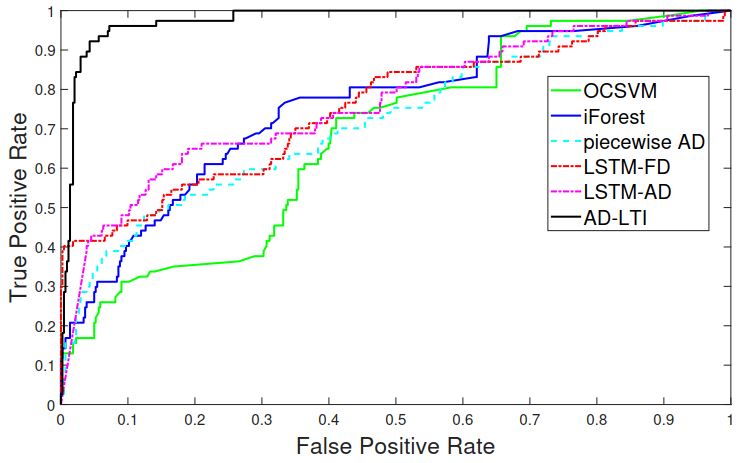}
    \caption{ROC curves of anomaly detection algorithms on Server Log dataset}
    \label{fig:ROC_serverlog}
\end{figure}

\begin{figure}[hbt]
    \centering
    \includegraphics[width=3.0in]{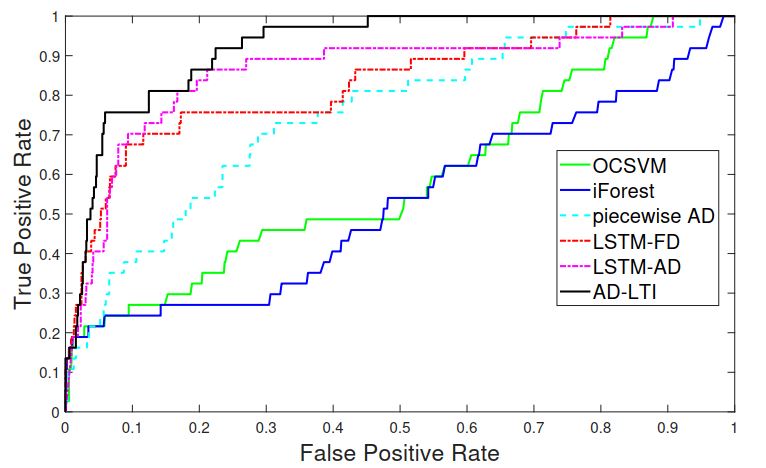}
    \caption{ROC curves of anomaly detection algorithms on Dodgers Loop dataset}
    \label{fig:ROC_dodgers}
\end{figure}

As shown in Fig. \ref{fig:heatmap_calit2}, Fig. \ref{fig:heatmap_serverlog} and Fig. \ref{fig:heatmap_dodgers}, we draw three groups of 1-D heatmaps to compare the detection decisions made by each algorithm (labelled on the y-axis) with the ground truth on each test dataset. Normal and anomalous frames are marked by green and red, respectively, on the map of ground truth. Frames are also marked by each anomaly detection algorithm with scores, which are reflected using a range of colors from green to red. Anomaly events are sparse in Calit2 dataset (Fig. \ref{fig:heatmap_calit2}) while comparatively more anomalous data point exist in the Server Log (Fig. \ref{fig:heatmap_serverlog}). From the figures, we can see that most of the ``hotspots" are captured by our scheme and its false alarm rate is comparatively low. Notably we also observe that OCSVM produces a large number of false alarms on Calit2 but fails to spot most of the anomaly frames on the Server Log dataset. The Piecewise method misses a lot of anomalies, while the iForest method tends to mistakenly label a large portion of normal data as anomalies. LSTM-AD produced the results close to our method on the Dodgers dataset, but rendered a large portion of false alarms on other two datasets. To give a more intuitive view, we plot the ROC curves of AD-LTI and the baseline algorithms on the test data in Fig. \ref{fig:ROC_calit2}, Fig. \ref{fig:ROC_serverlog} and Fig. \ref{fig:ROC_dodgers}.

From the ROC curves we can observe that AD-LTI produced the most reliable decisions as its curve is the closest to the top-left corner for all of the three datasets, especially on the Server Log Dataset (see Fig. \ref{fig:ROC_serverlog}), which features the complex seasonality in each channel. The detection difficulty on the Server Log dataset appears to be harder for other existing algorithms (the reason is explained later) - none of other algorithms achieve high true positive rate at a low false positive rate. We further calculate the corresponding AUC for each algorithm on both datasets. The resulting AUC values are shown in Table \ref{tab:AUC}.

\begin{table}[hbt]
\caption{Comparing the AUC values of anomaly detection algorithms on CalIt2, Server Log and Dodgers Loop datasets wherein actual contamination ratios (CR) are approximately 0.05, 0.15 and 0.10, respectively.}
\centering
\begin{tabular}{ l c c c } 
 \hline 
   		& CalIt2		& Server Log	& Dodgers Loop	\\
 \hline 
   OCSVM \cite{ocsvm}(default) 		& 0.876	& 0.677	& 0.591	\\
   OCSVM ($nu$ = CR) 				& 0.708	& 0.672	& 0.525	\\
   iForest \cite{isoforest}(default)& 0.891	& 0.756	& 0.535	\\
   iForest ($cr$ = CR)				& 0.877	& 0.761	& 0.518	\\
   Piecewise AD	\cite{piecewise}	& 0.833	& 0.721	& 0.751	\\
   LSTM-FD \cite{lstm-fd}			& 0.847	& 0.755	& 0.829	\\
   LSTM-AD \cite{lstm-ad}($L=L^*$)	& 0.900 & 0.793 & 0.859 \\
   AD-LTI ($L=L^*$)					& \textbf{0.935}	& \textbf{0.977}	&\textbf{0.923}\\
 \hline
\end{tabular}
\label{tab:AUC}
\end{table}

As shown in Table \ref{tab:AUC}, AD-LTI achieves the highest AUC values of 0.93, 0.977 and 0.923 on CalIt2, Server Log dataset and the Dodgers Loop datasets, respectively. On CalIt2, the AUC values of the baseline algorithms are between 0.8 and 0.9 with the only exception of OCSVM when $nu$ is set to 0.05 - the approximately actual anomaly rate (0.046, precisely) for CalIt2. This to some degree indicates that OCSVM is sensitive to parameters. Anomaly detection is much more challenging on the Server Log dataset due to the increase in the number of channels, and the complexity in seasonality and uncertainty (e.g., channel \emph{SU} is fairly unpredictable). As the result shows, the AUC values for all existing algorithms drop below 0.8 with the best of them, Isolation Forest, reaching 0.761 (with the contamination ratio $cr$ set to 0.15), which could float as it is a randomized algorithm. However, the actual contamination ratio is hardly a priori knowledge in practical scenarios. We also observed that prediction-driven approaches (LSTM-FD, LSTM-AD and AD-LTI) significantly outperformed others on the Dodgers Loop dataset -- this is mainly because of the presence of strong noise in the traffic data. The proposed AD-LTI algorithm makes the most reliable decisions in all of the tested scenarios. The main reasons are two-fold: from one perspective, the underlying backbone model for AD-LTI is very accurate with the complement of seasonal features that effectively captures complex seasonality and mitigates the noise in raw data. From another perspective, AD-LTI is robust in scoring each frame because we leverage multi-source forecasting and weight each prediction based on the confidence of the prediction source.

\begin{table*}[htb]
\centering
\caption{AUC values and detection overheads (in ms per frame) using LSTM-AD and AD-LTI under different settings of probe length $L$. Both methods use multiple forecasts with each frame being predicted for $L$ times.}
\begin{tabular}{l l c c c c c c} 
 \hline
 \multirow{2}{*}{Algorithm}&\multirow{2}{*}{$L$}&\multicolumn{2}{c}{Calit2 Dataset}&\multicolumn{2}{c}{Server Log Dataset}&\multicolumn{2}{c}{Dodgers Loop Dataset}\\ 
 &  & AUC & overhead(ms/frame) & AUC & overhead(ms/frame) & AUC & overhead(ms/frame)	\\
 \hline
 \multirow{4}{*}{LSTM-AD \cite{lstm-ad}}&$L = 5$ &0.900 &0.126 &0.793 &0.129 &0.859 &0.123\\
 &						  $L = 10$	&0.883	&0.142		&0.753 &0.148 	&0.778	&0.142 \\
 &						  $L = 20$	&0.847	&0.174		&0.596 &0.183 	&0.813	&0.174 \\
 &						  $L = 30$	&0.813	&0.206 		&0.505 &0.215 	&0.815	&0.205 \\
 \hline
 \multirow{4}{*}{AD-LTI} &$L = 5$	&0.911	&0.189		&0.977 &0.282 	&0.912	&0.196 \\
 &						  $L = 10$	&0.912	&0.353		&0.925 &0.399 	&0.923	&0.316 \\
 &						  $L = 20$	&0.935	&0.706		&0.845 &0.784 	&0.906	&0.707 \\
 &						  $L = 30$	&0.912	&1.125 		&0.784 &1.461 	&0.915	&1.110 \\ 
 \hline
\end{tabular}
\label{tab:varying_L}
\end{table*}

AD-LTI has an important hyper-parameter $L$, which determines both the prediction length for the backbone model and the maximum probe length for computing LTI. We evaluated our algorithm against LSTM-AD (which is also based on multiple forecasts) with different $L$ values to investigate the impact of $L$ on detection reliability and time efficiency. The result is summarized in Table \ref{tab:varying_L}.

From Table \ref{tab:varying_L} we can see our method outperformed LSTM-AD and also observe different impacts of the probe window length \textit{L} on different datasets. On CalIt2 and Dodgers, the impact of \textit{L} on the detection reliability (revealed by AUC) is subtle, while on the Server Log dataset very large $L$ values show obvious negative effect on our scheme. The reasons behind these results are partly because as $L$ becomes bigger ($L$ is set to 20 or above), the prediction made by the backbone model becomes less accurate, and partly because of the dilution of local information. In comparison, LSTM-AD is much more susceptible to the hyper-parameter $L$. Besides, as expected a longer probe length leads to the increased overhead in detection, which can be mitigated by running the scheme in parallel. Empirically, we recommend setting $L$ to a value between 5 and 20 considering both detection reliability and efficiency.

\section{Conclusion}
On-line detection of anomalies in time series has been crucial in a broad range of information and control systems that are sensitive to unexpected events. In this paper, we propose an unsupervised, prediction-driven approach to reliably detecting anomalies in time series with complex seasonality. We first present our backbone prediction model, which is composed of a time series decomposition module for seasonal feature extraction, and an inference module implemented using a GRU network. Then we define Local Trend Inconsistency, a novel metric that measures abnormality by weighting local expectations from previous records. We then use a scoring function along with a detection algorithm to convert the $LTI$ value into the probability that indicates a record's likelihood of being anomalous. The whole process can leverage the matrix operations for parallelization. We evaluated the proposed detection algorithm on three different datasets. The result shows that our scheme outperformed several representative anomaly detection schemes commonly used in practice.

In the future we plan to focus on extending our work to address new challenges in large-scale, information-intensive distributed systems such as edge computing and IoT. We aim to refine our method with scenario-oriented designs, for instance, detection in asynchronized streams sent by distributed sensors, and build a robust monitoring mechanism in order to support intelligent decisioning in these types of systems.

\section*{Acknowledgement}
This work is partially supported by Worldwide Byte Security Co. LTD, and is supported by National Natural Science Foundation of China (Grant Nos. 61772205, 61872084), Guangdong Science and Technology Department (Grant No. 2017B010126002), Guangzhou Science and Technology Program key projects (Grant Nos. 201802010010, 201807010052, 201902010040 and 201907010001), and the Fundamental Research Funds for the Central Universities, SCUT (Grant No. 2019ZD26).

\end{document}